\documentclass[journal]{IEEEtran}
\usepackage{amsmath,amssymb,amsfonts}
\usepackage{amsopn}
\usepackage{graphicx}
\usepackage{subfigure}
\usepackage{epstopdf}
\usepackage{multirow}
\usepackage{array}
\usepackage{algorithm}
\usepackage{algorithmic}
\usepackage{arydshln}
\usepackage{amsthm,bm}
\usepackage{expl3}
\ExplSyntaxOn
\newcommand\latinabbrev[1]{
  \peek_meaning:NTF . {
    #1\@}%
  { \peek_catcode:NTF a {
      #1.\@ }%
    {#1.\@}}}
\ExplSyntaxOff

\def\etal{\latinabbrev{et al}}

\def\ie{\latinabbrev{i.e}}

\ifCLASSINFOpdf
\else
\fi

\begin{document}
%
\title{Regression-based Hypergraph Learning for Image Clustering and Classification}
%
%
%

\author{Sheng~Huang~\IEEEmembership{Student~Member,~IEEE},
        Dan~Yang,
        Bo~Liu,
        Xiaohong~Zhang
\thanks{Sheng Huang, Dan Yang and Xiaohong Zhang are with the School of Software Engineering, Chongqing University, Chongqing, 40004 PRC Email: \{huangsheng, dyang, xhongz\}@cqu.edu.cn}
\thanks{Bo Liu is with the Department of Computer Science, Rutgers University, Piscataway, NJ, 08854 USA Email: lb507@rutgers.edu}}


\markboth{IEEE Transactions on }%
{Shell \MakeLowercase{\textit{et al.}}: Bare Demo of IEEEtran.cls for Journals}

\maketitle

\begin{abstract}
Inspired by the recently remarkable successes of Sparse Representation (SR), Collaborative Representation (CR) and sparse graph, we present a novel hypergraph model named Regression-based Hypergraph (RH) which utilizes the regression models to construct the high quality hypergraphs. Moreover, we plug RH into two conventional hypergraph learning frameworks, namely hypergraph spectral clustering and hypergraph transduction, to present Regression-based Hypergraph Spectral Clustering (RHSC) and Regression-based Hypergraph Transduction (RHT) models for addressing the image clustering and classification issues. Sparse Representation and Collaborative Representation are employed to instantiate two RH instances and their RHSC and RHT algorithms. The experimental results on six popular image databases demonstrate that the proposed RH learning algorithms achieve promising image clustering and classification performances, and also validate that RH can inherit the desirable properties from both hypergraph models and regression models.
\end{abstract}


\IEEEpeerreviewmaketitle

\section{Introduction}
As the generalization of the graph model~\cite{hyper,expansion}, the hypergraph model is more flexible and more intuitive to depict the complex relation of data, since the edge of hypergraph, which is known as the hyperedge, can contain more than two vertices. Due to this desirable property, hypergraph learning is recently drawn intensive attention. Over past decades, extensive hypergraph learning approaches has been proposed and successfully applied to tackle a lot of fundamental tasks, such as clustering~\cite{hyper,hyperweight}, classification~\cite{adaptive,l1h,hypersvm}, segmentation~\cite{higher}, dimensionality reduction~\cite{dhlp,grlda} and multi-label learning~\cite{mlc,hap}.

As same as graph learning, the hypergraph construction process plays a vital role in hypergraph learning and a good quality hypergraph should well reveal the real relation of samples. Hypergraph learning is a frequently used tool for unsupervised and semi-supervised learning. In these two cases, the previous hypergraph learning works often adopt the neighbourhood-based (distance-based) strategy to build the hypergraph~\cite{hyper,phr,sum}. More specifically, for each sample, a hyperedge is generated by connecting this centroid sample and its $k$ nearest neighbors. However, as such neighborhood-based often cannot well even correctly discover the real relations of samples and it is also very sensitive to noises, the quality of hypergraph is lowered which directly degrades the performance of hypergraph learning.

\begin{figure*}[!tbp]
\centering
\subfigure[Raw Samples and their rank scores]{
\centering
\includegraphics[scale=0.45]{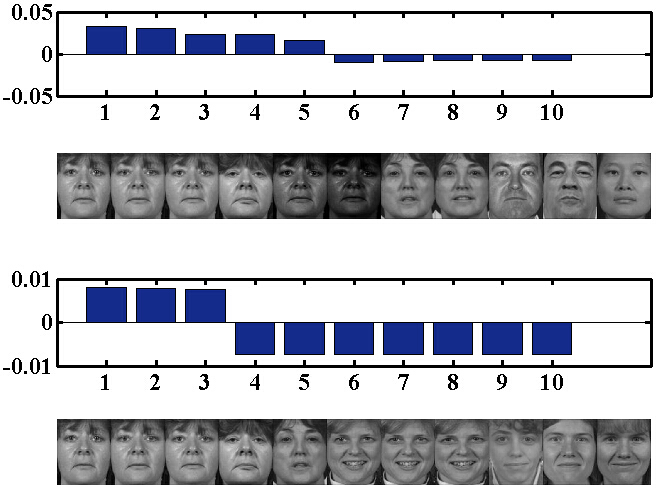}
\label{}}
\subfigure[Samples with noise and their rank scores]{
\centering
\includegraphics[scale=0.45]{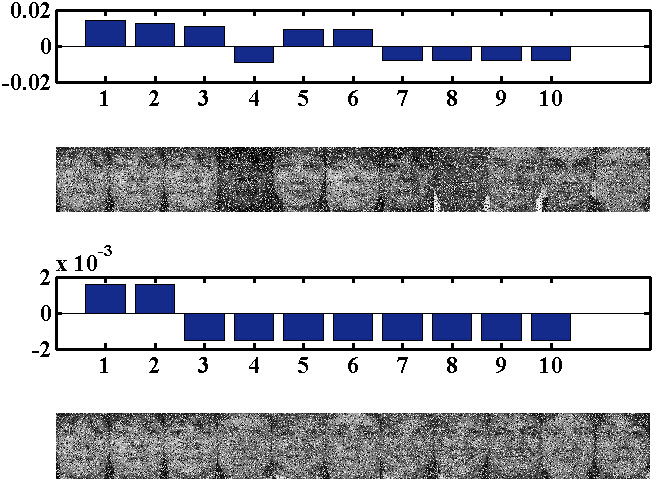}
}
\caption{This figure was originally shown in our early work~\cite{sgc}. We use it here for intuitively specifying the advantages of SR over the conventional neighbourhood-based method in the relevant sample selection procedure. The figure shows the top 10 most relevant face images selected by SR and K-Nearest Neighbour (KNN) based on a given query face image. This experiment is conducted in a subset of FERET database~\cite{feret} (72 subjects with 6 images in each subject). \textbf{The first two rows of the figure are the selection results of SR while the last two rows are the selection results of KNN.} The left subfigure reports the results on the original FERET database while the right one reports the results on the modified FERET database in which 30\% of pixels of each image has been corrupted by noise. In the figure, the first face image of each image array is the query image and the rest ten images are the relevant face images selected by SR or KNN. The histograms above the image array demonstrates the confidence scores of these top ten relevant face images. If the subjects of the return face image and the query face image are identical,  its corresponding histogram is positive otherwise it is negative.  In the figure, SR gets five hits either on the original FERET database or on the noisy FERET database while KNN only gets three and two hits on these two datasets respectively. Clearly, this phenomenon demonstrates the advantages of SR in relevant sample selection.}
\label{Examples}
\end{figure*}

In the recent decade, Sparse Representation (SR) has achieved remarkable successes for addressing dozens of computer vision and machine learning issues~\cite{src,ksrc,wsrc}. The main merits of SR are the strong discriminating power and the excellent robustness to noises which endow SR with a better related sample selection capacity for a given sample in comparison with the conventional neighborhood-based approaches (We refer to an toy example in our early work~\cite{sgc} to experimentally verify this argument, please see Fig~\ref{Examples}). In other words, SR can better discover the real relations of data. Motivated by this fact, several novel graph learning approaches have been developed via leveraging SR to construct the graphs~\cite{l1graph,sgc,srdp,spp,ssc}. Compared to the traditional graph learning approaches, these works have achieved better performances. Since the hypergraph model is the generalization of the graph model and hypergraph learning is closely related to graph learning, we believe that the success of SR in graph learning should also be applied to hypergraph learning. Moreover, SR is essentially a $L_1$ or $L_0$-norm regularized regression model and its success also motivates the presentations of many influential regression models which often enjoy some desirable properties~\cite{src,ksrc,crc,rcr,wcrc,cssr}. Cleary, these successful works can also provide some new ways for graph or hypergraph construction.

In this paper, we generalize the idea of sparse graph to present a novel hypergraph construction framework, which can leverage the regression model to construct the high quality hypergraph. We name such framework Regression-based Hypergraph (RH) model. More specifically, in RH, each sample represents a vertex and constructs a regression system together with the rest samples for measuring correlations of samples. Then, based on the obtained correlations, each sample and its top $m$ most relevant samples are employed to define a hyperedge. Moreover, the mean of the correlations among samples in a hyperedge is considered as the weight of this hyperedge, since the correlation is an intuitive measure of the closeness between two samples. We also plug the regression-based hypergraph into two classical hypergraph learning frameworks, namely hypergraph spectral clustering and hypergraph transduction, to present Regression-based Hypergraph Spectral Clustering (RHSC) and Regression-based Hypergraph Transduction (RHT) models for addressing the clustering and classification issues. As two of the most influential regression models for visual learning, Sparse Representation (SR) and Collaborative Representation (CR) are adopted as two examples to instantiate two RH instances. Since SR and CR are actually the $L_1$-norm and $L_2$-norm regularized regression models, we name these two instances $L_1$-Hypergraph ($L_1$H) and $L_2$-Hypergraph ($L_2$H) respectively. Similarly, their hypergraph spectral clustering and hypergraph transduction algorithms are named as $L_n$-Hypergraph Spectral Clustering ($L_n$HSC) and $L_n$-Hypergraph Transduction ($L_n$HT) where $n = 1$ or 2 if $L_1$H or $L_2$H is applied.

Regression-based Hypergraph (RH) model inherits the advantages of both hypergraph and regression models. Compared to the conventional hypergraph models, RH can incorporate some properties from the chosen regression models. For example, if the Sparse Representation (SR) or Collaborative Representation (CR) is selected for hypergraph construction, the constructed RH should be more discriminative and robust, since these two regression models are better to discover the relevances among samples over the conventional neighborhood-based hypergraph construction fashions. Compared to the regression approaches,  RH constructs a hypergraph to sufficiently exploit the correlation of each pair of samples instead of just utilizing the correlations between the target sample and the other samples as the regression approaches do. Compared to the regression-based graph approaches, such as the sparse graph, RH is a hypergraph model which owns a better capability and flexibility to depict the complex high-order data relations.

We employ six popular visual databases to validate our works. The experimental results demonstrate the superiority of the RH model over the conventional hypergraph models. We conclude three main contributions of our works as follows:
\begin{enumerate}
 \setlength{\itemsep}{0pt}
 \setlength{\parskip}{0pt}
 \setlength{\parsep}{0pt}
 \item We provide a general idea which utilizes the regression models to construct the high quality hypergraphs. To the best of our knowledge, this paper is the first to formally and systematically build a bridge between the hypergraph model and the regression model.
 \item We present two novel hypergraph learning frameworks based on the RH model to tackle the clustering and classification tasks respectively.
 \item We adopt two recently influential regression models, namely Sparse Representation (SR) and Collaborative Representation (SR), to instantiate two RH instances called $L_1$-Hypergraph ($L_1$H) and $L_2$-Hypergraph ($L_2$H) which are experimentally proved to be more discriminative and robust than the conventional hypergraphs.
\end{enumerate}

The rest of paper is organized as follows: the previous works are reviewed in Section~\ref{related}. Section~\ref{method} introduces the methodology of our works; experiments are presented in section~\ref{exp}; the conclusion is finally summarized in section~\ref{conclude}.

\section{Previous Works}
\label{related}
\subsection{Regression Models}
Regression model is a common technique for data analysis and has been successfully applied to almost all the areas in computer vision, machine learning and image processing~\cite{lrc,src,crc}. Sparse Representation (SR) may be the most influential regression approach in the recent decade. SR is mainly inspired by the idea of compressed sensing~\cite{cs}. In SR, a $L_0$ or $L_1$-norm constraint is introduced to the common regression model for compulsively selecting only a few of relevant measurements and ignoring the irrelevant ones by assigning their corresponding regression coefficients to zero. This endows SR with a strong discriminating power and a good robustness. However, Zhang \etal~\cite{crc,wcrc} argued that the collaboration of samples instead of the sparsity is the essential factor that leads to such good discriminating ability and robustness. They proposed a linear regression model named Collaborative Representation (CR) via employing a relatively mild $L_2$-norm constraint to replace the $L_1$-norm constraint to achieve the collaboration property. Many works have shown that CR is more efficient and can get a similar or even better performance. Due to the desirable properties of SR and CR, they have achieved remarkable successes in many areas and promotes the presentations of many impressive regression approaches for addressing different computer vision, machine learning and image processing issues. For examples, Gao \etal~kernelized SR for face recognition and image classification~\cite{ksrc}. Yuan \etal~presented a multitask joint sparse representation model to combine the strength of multiple features and/or instances for visual classification~\cite{mjsr}. Huang \etal~presented a SR-based classifier named Class Specific Sparse Representation (CSSR) which incorporated the properties of both SR and CR~\cite{cssr} via defining the homogenous samples as a group and making them competition for representing the test sample. Yang \etal~proposed Relaxed Collaborative Representation (RCR) to effectively exploit the similarity and distinctiveness of samples~\cite{rcr}. Although these regression approaches have obtained promising performances in different fields, they all have an obvious drawback that they can only utilize the correlation between the testing sample and the training samples. On the contrary, the proposed Regression-based Hypergraph (RH) model can sufficiently exploit the correlations among all samples. Another merit of RH is that there exists extensive regression approaches which can bring more flexility to the hypergraph model.

\subsection{Sparse Graph}
Since Sparse Representation (SR) is good at selecting the relevant samples for a test sample even in the noisy conditions, some researchers have attempted to use SR to construct high quality graphs for addressing different issues. In these works, such constructed graphs are often called $L_1$-graph or sparse graph and have achieved very promising performances. More specifically, Qiao \etal~and Timofte \etal~successively use SR to construct a sparse graph for dimensionality reduction~\cite{srdp,spp}. Huang \etal~leverage SR to measure the correlations between each two samples and then construct a sparse graph for transduction~\cite{sgt}. The Sparse Subspace Clustering (SSC) algorithms~\cite{ssc,sssc,rss} learn a sparse graph for clustering via considering the data self representation problem as a SR issue. Similar to~\cite{srdp,spp}, Cheng \etal~utilize SR to construct the $L_1$-graph (sparse-graph) for spectral clustering, subspace learning and semi-supervised learning~\cite{sparsegraph}.  Although the applications and the learning (or construction) procedures of these works are very different, the obtained sparse graphs are very similar which all demonstrate the better discriminative abilities and robustness over the conventional graph models. The main drawback of the sparse graph models is that they cannot intuitively describe the high-order complex data relations, because these sparse graph models are essentially graph model whose edges can only depict the simple pairwise data relation. Since Regression-based Hypergraph (RH) model is deemed as a generalization of sparse graph from the perspectives of both regression and hypergraph, it does not suffer from this issue.

\subsection{Hypergraph Models}
As a generalization of graph, hypergraph represents the structure of data via measuring the similarity between groups of points~\cite{sum,phr,adaptive,he,hyper,smhyper}. The main difference between graph and hypergraph is that the edge of hypergraph can own more than two vertices which endows hypergraph with a high flexility for depicting the high-order relation. Benefitted by this desirable property, hypergraph models have been successfully applied into dozens of computer vision, machine learning and pattern recognition areas. In the past, the researchers were more keen to develop different hypergraph frameworks which define different theories to depict the hypergraph structure. The representative approaches include Clique Expansion~\cite{expansion}, Star Expansion~\cite{expansion}, Zhou's Normalized Laplacian~\cite{hyper}, Clique Averaging~\cite{mean}, Bolla's Laplacian~\cite{bolla} and so on. However, as was shown in \cite{hol}, all of the previous approaches, despite their very different formulations, can be proved to be equivalent to each other under specific conditions. Currently, the researchers pay more attention on developing the algorithms for the hypergraph constructions under the aforementioned hypergraph frameworks. In hypergraph, hyperedge defines the relation of data. Therefore, the hyperedge generation is very crucial to the quality of the constructed hypergraph. Conventionally, most of hypergraph models adopt the neighbourhood-based fashion to generate the hyperedges. For examples, Huang \etal~proposed a hypergraph learning framework for image retrieval, in which each image and its $k$-nearest neighbors form the hyperedge~\cite{phr}. Zhou \etal~also adopted such neighbourhood-based fashion to generate the hyperedges for unsupervised and semi-supervised hypergraph learning~\cite{hyper}. The main problems of these approaches are that they often cannot well reveal the real relation of data and are sensitive to noise. Some researchers also employed the clustering techniques to generate the hyperedges and then construct the hypergraph. As the representative approach of such category, Gao \etal~proposed a hypergraph-based 3-D object retrieval approach via utilizing the $k$-means to cluster the views of the 3-D objects and consider each cluster as a hyperedge~\cite{sum}. Since the hyperedges, which are formed by the clusters, cannot share intersection vertices, these hypergraphs cannot capture the correlations of data. Another popular method is to adaptively contruct the hypergraph via imposing some meaningful constraints. As an instance of this category, Yu \etal~introduced a $L_2$-norm constraint to the hyperedge weight matrix to present a hypergraph transduction approach for image classification~\cite{adaptive}. This method generates the hyperedges via adaptively assigning the weights to the hyperedges. However, it cannot guarantee the inexistence of the isolated vertices. Similar to the work~\cite{adaptive}, Wang \etal~imposed a Laplacian cost constraint and a $L_1$-norm constraint to the hyperedge weights for adaptively learning the hyperedge weights in a hypergraph model~\cite{l1h}. In its hyperedge generation procedure, the traditional neighbourhood-based fashion is employed to define a candidate hyperedge vertex set and then SR is applied to prune noisy vertices in this set for forming the final hyperedge. Such idea is similar but also different to us. It still considers the neighbours of a sample as its relevant samples and SR here only plays a role as a noise remover. On the contrary, Regression-based Hypergraph (RH) model thoroughly utilizes the regression model (includes SR) to generate the hyperedged. And RH is more formal and systematic to introduce how to use regression models to construct the high quality hypergraph.



\section{Methodology}
\label{method}
\subsection{Regression-based Hypergraph}
In order to incorporate some desirable properties of the regression algorithms, we introduce a new hypergraph learning framework named Regression-based Hypergraph (RH), which leverages different regression models to construct the hypergraphs. Let a $d\times n$-dimensional matrix $X=[x_1,x_2,\cdots,x_n]$ be the sample matrix, where $d$ is the dimension of sample and $n$ is the number of sample. The $d$-dimensional column vector $x_i$ is a sample which is also the $i$-th column of sample matrix. We apply the general formulation of regression model to estimate the correlations between each sample and the rest samples,
\begin{equation}\label{regression}
  \hat{C}_i=\arg\underset{C_i}\min~{\Delta(x_i,X_{t\neq i}C_i)+\beta\Phi(C_i)}
\end{equation}
where $\Delta(\cdot)$ and $\Phi(\cdot)$ are the regression error and the regularization term respectively. $X_{t\neq i}$ is the sample matrix which excludes the $i$-th sample $x_i$. The ($n-1$)-dimensional column vector $C_i=[c_{i1},\cdots,c_{i(i-1)},c_{i(i+1)},\cdots,c_{in}]^T$ is the regression coefficient vector with respect to the sample $x_i$. Each element of regression coefficient vector encodes the correlation between the target sample and the sample. According to the aforementioned correlation computation fashion, each pair of samples can get two correlations, \ie, the samples $x_i$ and $x_j$ has two correlations $c_{ij}$ and $c_{ji}$. Extensive literatures~\cite{l1graph,srdp,sslsr,sgt,nsgssl} show that such correlation between two samples is a high quality similarity measure of samples. Therefore, following the sample similarity computation fashion in~\cite{l1graph,srdp},  we define the sample similarity as the mean of the correlation absolute values of each pair of samples to guarantee the nonnegativity and symmetry of the similarity. More specifically, the similarity between the sample $x_i$ and the sample $x_j$ can be mathematically denoted as follows
\begin{equation}\label{}
  s_{ij}=s_{ji}=\frac{|c_{ij}|+|c_{ji}|}{2}.
\end{equation}
The regression model cannot compute the self-correlations of samples. In other words, the self-similarity computation of a sample is still not provided. In such case, we define the self-similarity of a sample as the sum of the similarities between this sample and the rest samples, $s_{ii}=\sum_{t\neq i}s_{it}$. According to the obtained similarities among all the samples, it is not hard to construct a similarity matrix (or affinity matrix) $S$ where $s_{ij}$ is the $ij$-th element of $S$. Then, the normalization of the sample similarities can be done as follows
\begin{equation}\label{}
  S=\sqrt{M}S\sqrt{M}
\end{equation}
where $n\times n$-dimensional matrix $M$ is diagonal matrix whose $i$-th diagonal element is the sum of elements in the $i$-th row of $S$, $m_{ii}=\sum_{t}s_{it}$.

After obtaining the similarities, we define a hypergraph $G(V,E)$ for depicting the relations among samples where $V$ and $E$ are the collections of its vertices and hyperedges respectively. In this hypergraph, each sample is deemed as a vertex, \ie, the sample $x_i$ is corresponding to the vertex $v_i\in V$. Same as the conventional hypergraph construction fashion, a sample $x_i$ and its top $t-1$ most similar samples are employed to define a $t$-length hyperedge $e_i\in E$. Therefore, for a data collection constructed by $n$ samples, we can obtain $n$ hyperedges. The weight of the hyperedge $e_i$ is defined as the mean similarity of samples in this hyperedge,
\begin{equation}\label{}
  w_i=\frac{\sum_{\{v_a,v_b\}\in e_i}s_{ab}}{l_i}
\end{equation}
where $l_i$ is the number of pairs of vertices in the hyperedge $e_i$.

\subsection{Learning with Regression-based Hypergraph}
Spectral clustering and hypergraph transduction are the most common unsupervised and supervised hypergraph learning techniques respectively. In this subsection, we apply our Regression-based Hypergraph (RH) model to these two techniques for validating the effectiveness of our model.
We develop a novel spectral clustering and hypergraph transduction framework and name them Regression-based Hypergraph Spectral Clustering (RHSC) and Regression-based Hypergraph Transduction (RHT) respectively. We begin by introducing some common definitions of hypergraph learning~\cite{hyper}. We denote the degrees of vertex and hyperedge as the sum of weights of hyperedges which are incident to the given vertex and the number of vertices in the hyperedge respectively. Mathematically, the degrees of vertex and hyperedge are respectively reformulated as $d(v_i) = \sum_{\{v_i\in e_j| e_j\in E\}} w_j$ and $\delta(e_i) =|e_i|$. The vertex-edge incident matrix $H$ is a common tool for depicting the structure of hypergraph. each of its rows and columns are corresponding to the vertex and the hyperedge of hypergraph respectively. More specifically, for a hypergraph consisted by $n$ vertices and $m$ hyperedges, its vertex-edge incident matrix is a $n\times m$-dimensional binary matrix. If the vertex $v_i$ is on the hyperedge $e_j$, the ($i,j$)-th element of $H$ is 1, otherwise, 0. Due to the hyperedge generation fashion of RH, the dimension of its vertex-edge incident matrix is $n\times n$.

From the perspective of graph learning, the spectral clustering is actually a graph (or hypergraph) partition issue~\cite{hyper,sc,ncut}. Then, we can consider the regression hypergraph-based spectral clustering problem as a normalize hypergraph cut issue. According to Zhou's work~\cite{hyper}, such issue can be solved by following optimization model,
\begin{equation}\label{hcut}
\small
    \hat{F}=\arg\underset{F}\min~\Omega(F, G):=\frac{1}{2}\sum_{e_i\in E}\sum_{(v,u)\in e_i}\frac{w_i}{\delta(e_i)}\left|\left|\frac{F(u)}{\sqrt{d(u)}}-\frac{F(v)}{\sqrt{d(v)}}\right|\right|^2,\\
\end{equation}
where the $n\times m$-dimensional matrix $F=[f_1,\cdots,f_m]$ is the collection of the hypergraph cuts of the given regression hypergraph $G(V,E)$. The $n$-dimensional column vector $f$ is a hypergraph cut which introduces a binary partition to the given hypergraph and its elements indicate the confidences of how the corresponding vertices belonging to a subgraph after partition. $F(v)$ is a row of matrix $F$ which encodes the elements of hypergraph cuts corresponding to the vertex $v$.

According to the normalized cut criterion~\cite{ncut}, the optimal hypergraph cuts should maximize the compactness of partitioned subgraphs and minimize the compactness of the boundaries between the subgraphs simultaneously. The compactness of subgraphs and boundaries is measured by the normalized summation of the hyperedge weights of a vertex set. With several reductions, Equation~\ref{hcut} can be further translated into the following matrix expression,
\begin{eqnarray}\label{dhcut}
 \nonumber
    \hat{F}&=&\arg\underset{F}\min~\Omega(F, G)\\
    &:=&\frac{1}{2}\sum_{e_i\in E}\sum_{(v,u)\in e_i}\frac{w_i}{\delta(e_i)}\left|\left|\frac{F(u)}{\sqrt{d(u)}}-\frac{F(v)}{\sqrt{d(v)}}\right|\right|^2\\\nonumber
  &:=&\text{Trace}\{F^T(I-D_v^{-1/2}{HW D_e^{-1}H^T}{D_v^{-1/2}})F\}\\ \nonumber
  &:=&\text{Trace}(F^TL_RF),
\end{eqnarray}
where function $\text{Trace}(\cdot)$ returns the trace of matrix. $D_{e_i}$, $D_v$ and $W$ are the diagonal matrix forms of $\delta(e_i)$, $d(v)$ and $w_i$ respectively. $I$ is the identity matrix and $L_R$ is the derived normalized hypergraph Laplacian matrix which encodes the structure of the regression hypergraph $G$. The detail deductions of this equation can be referred to the works~\cite{hyper,hap}.

The problem in Equation~\ref{dhcut} is a typical eigenvalue problem. It can be easily solved by eigenvalue decomposition technique. The top $m$ optimal hypergraph cuts $F=[f_1,\cdots,f_m]$ are exactly the top $m$ eigenvectors corresponding the top $m$ minimal nonzero eigenvalues. Finally, the learned hypergraph cut collection $F$ is deemed as the new representation of data for clustering.

Graph-based transduction is a semi-supervised learning technique which is often leveraged to address the labeling and classification issues. In the semi-supervised case, the labels of some of data are available. Therefore, an optimal hypergraph cut should not only minimize the loss of the geometric structure of data (the loss of data relation) but also minimize the labeling error. Conventionally, the labeling error is measured by the Euclidean distance between the labels and the hypergraph cuts, since the hypergraph cuts can be deemed as the collection of the label indicator of vertex. Thus, the original hypergraph partition model in Equation~\ref{hcut} can be further improved as the following regularized hypergraph partition model for considering the labeling error of data,
\begin{eqnarray}\label{predmodel}\nonumber
  \hat{F}&=&\arg\underset{F}\min\{\Omega(F, G)+\lambda \Delta(F, Y)\}\\
  &=&\text{Trace}(F^TL_RF)+\lambda ||F-Y||^2
\end{eqnarray}
where $\lambda$ is a positive parameter to reconcile these two losses and the $n\times c$-dimensional matrix $Y=[y_1,\cdots,y_c]$ is the collection of labels. $y_i$ is the label vector of the $i$-th class. Let us denote $y_i(v)$ is the label of vertex $v$. Then, we have $y_i(v)=1$ or -1 if the vertex $v$ belonging to $i$-th class or other classes respectively, and 0 if the vertex $v$ is unlabeled. Note, here the collection of hypergraph cuts $F$ has the same size as $Y$ where $m=c$.

Such problem is a typical Least Square (LS) problem which can be efficiently solved. According to works~\cite{hyper,adaptive}, its solution is as follows
\begin{equation}\label{}
 F=\frac{1}{1+\lambda} {\left( \frac{L_R+\lambda I}{1+\lambda}\right)^{-1}} Y.
\end{equation}
After obtaining $F$, the labeling (or classification) of $i$-th sample can be accomplished by assigning it to the $t$-th class that satisfies $t=argmax_j F_{ij}$.

\subsection{Two RH Instances: $L_1$-Hypergraph and $L_2$-Hypergraph}
Sparse Representation (SR) and Collaborative Representation are two of the recent most influential regression models in computer vision and machine learning. We employ them to instantiate two Regression-based Hypergraph (RH) instances. Since SR and CR are actually the $L_1$-norm and $L_2$-norm regularized regression models, we name these two RH instances $L_1$-Hypergraph ($L_1$H) and $L_2$-Hypergraph ($L_2$H) respectively. The detail information of $L_1$H and $L_2$H are presented in Table~\ref{regression}. We have also plug $L_1$H and $L_2$H into Regression-based Hypergraph Transduction (RHT) and Regression-based Hypergraph Spectral Clustering (RHSC) frameworks to produce four new semi-supervised or unsupervised hypergraph learning approaches respectively named $L_1$-Hypergraph Transduction ($L_1$HT), $L_2$-Hypergraph Transduction ($L_2$HT), $L_1$-Hypergraph Spectral Clustering ($L_1$HSC), and $L_2$-Hypergraph Spectral Clustering ($L_2$HSC). The relations of these approaches are shown in Table~\ref{relation}. In the next section, we apply these four RH instance algorithms to demonstrate the superiorities of our models over the conventional hypergraph models and verify the assumption that RH should inherit some desirable properties from the chosen regression models.
\begin{table}[!tbp]
\renewcommand{\arraystretch}{1.3}
    \caption{The detail information of two mentioned RH instances.}
    \label{instance}
    \vspace{-0.3cm}
\begin{center}
    \begin{tabular}{c | c |c |c}
    \hline
     Name& Regression model& $\Delta(\cdot)$ in Equation~\ref{regression}& $\Phi(\cdot)$ in Equation~\ref{regression}\\
     \hline
    $L_1$H&SR~\cite{src}&$||x_i-X_{t\neq i}C_i||_2^2$&$||C_i||_1$\\
    $L_2$H&CR~\cite{crc}&$||x_i-X_{t\neq i}C_i||_2^2$&$||C_i||_2^2$\\
    \hline
    \end{tabular}
\end{center}
\end{table}

\begin{table}[!tbp]
\renewcommand{\arraystretch}{1.3}
    \caption{The relations of the proposed four hypergraph learning approaches.}
    \label{relation}
    \vspace{-0.3cm}
\begin{center}
    \begin{tabular}{c | c c }
    \hline
     \multirow{2}*{RH Instances}
    &\multicolumn{2}{c}{Hypergraph Learning Frameworks}\\ \cline{2-3}
    & RHT (Semi-Supervised) & RHSC (Unsupervised) \\
     \hline
    $L_1$H (SR~\cite{src})&$L_1$HT&$L_1$HSC\\
    $L_2$H (CR~\cite{crc})&$L_2$HT&$L_2$HSC\\
    \hline
    \end{tabular}
\end{center}
\end{table}


\section{Experiments}
\label{exp}
In this section, we conduct some experiments to employ the aforementioned four RHSC and RHT intances, namely $L_1$HSC, $L_2$HSC, $L_1$HT and $L_2$HT, to tackle the image clustering and classification tasks respectively. Since these four algorithms are all generated from Sparse Representation (SR) or Collaborative Representation (CR) which enjoy the robustness to noise and occlusion, we also conduct some experiments to discuss if these four algorithms have inherited such desirable property.
\begin{figure}[!tb]
\centering
\subfigure[AR]{
\centering
\includegraphics[scale=0.57]{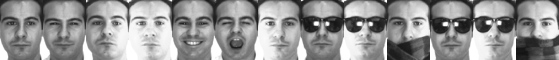}
}
\subfigure[ORL]{
\centering
\includegraphics[scale=0.26]{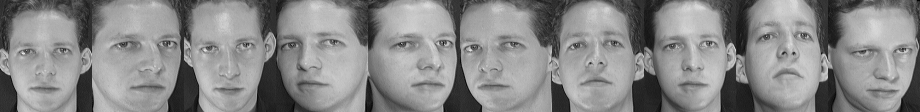}
}
\subfigure[COIL20]{
\centering
\includegraphics[scale=0.56]{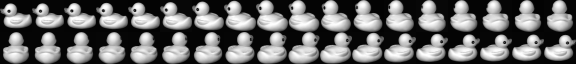}
}
\subfigure[ETH80]{
\centering
\includegraphics[scale=0.56]{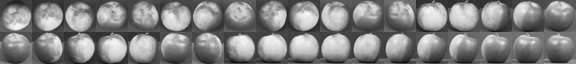}
}
\subfigure[Scene15]{
\centering
\includegraphics[scale=0.34]{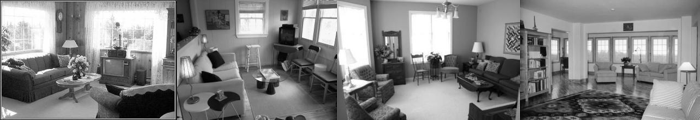}
}
\subfigure[Caltech256]{
\centering
\includegraphics[scale=0.34]{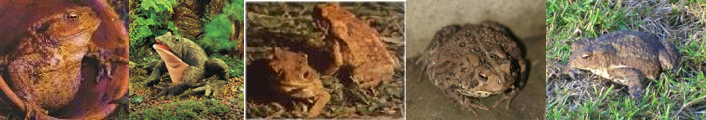}
}
\caption{The samples from six involved image databases.}
\label{examples}
\end{figure}

\subsection{Datasets and Compared Methods}
Six image datasets, named AR~\cite{AR}, ORL~\cite{orl}, COIL20~\cite{coil}, ETH80~\cite{eth80}, Scene15~\cite{scene15} and Caltech256~\cite{caltech256}, are leveraged for validating our works. AR face database consists of more than 4,000 color images of 126 subjects~\cite{AR}. Following paper~\cite{lrc,dhlp}, a subset contains 2600 images with 100 subjects are constructed in our experiment. Each subject has 26 images. The first 14 images of each subject are not involved any occlusion while the rest 12 images are involved the occlusions. In these 12 images, the faces in the first six images are occluded by the sunglasses and the faces in the other six images are occluded by the scarfs. In the general image classification and clustering experiments, only the images without any occlusion are utilized. The whole dataset is leveraged to analysis the robustness of the proposed work to disguise. The size of face image on AR database is 60$\times$43 pixels. ORL database is a face image database, which contains 400 images from 40 subjects~\cite{orl}. Each subject has ten images acquired at different times. The size of face image on ORL database is 32$\times$32 pixels. The COIL-20 database has 20 objects and each object has 72 images which are obtained by the rotation of the object through 360$^\circ$ in 5$^\circ$ steps (1440 images in total) \cite{coil}. The size of each image is 32$\times$32 pixels on COIL20 database. The ETH80 object database~\cite{eth80} contains 80 objects from 8 categories. Each object is represented by 41 views spaced evenly over the upper viewing hemisphere (3280 images in total). The original size of each image in this dataset is 128$\times$128 pixels. We resize them to 32$\times$32 pixels. Scene15 database~\cite{scene15} is a scene database, which has 15 classes with 100 samples per category. Following paper~\cite{hyper}, a subset of Caltech256 database~\cite{caltech256}, which has 20 classes with 100 samples per category, is used in our experiments. We directly use grayscale as the feature on ORL, AR, COIL20 and ETH80 databases. PiCoDes~\cite{picodes} is adoptted to represent the images on Scene15 and Caltech256 databases, since they are more challenging. The dimension of PiCoDes feature is 2048. Figure~\ref{examples} shows some samples of these six image databases.

Nonnegative Matrix Factorization (NMF)~\cite{nmf}, Graph regularized Nonnegative Matrix Factorization (GNMF)~\cite{gnmf}, Normalized Cut (NCut)~\cite{ncut}, Normalized Hypergraph Spectral Clustering (NHSC)~\cite{hyper}, Large-scale Spectral Clustering (LSC)~\cite{lsc}, Clique Expansion-based Hypergraph Spectral Clustering using Matrix Trace Weights (CEHSC+Trace)~\cite{hyperweight}, Normalized Hypergraph Spectral Clustering using the mean of distances between the centroid and the vertices in a hyperedge (NHSC+Cent)~\cite{hyperweight}, and $L_1$-Graph Spectral Clustering ($L_1$GSC)~\cite{l1graph} are chosen as the compared approaches in the image clustering experiments. NCut can be deemed as a sort of regular graph spectral clustering algorithm. So there are three graph spectral clustering algorithms. They are Ncut, LSC and $L_1$GSC. NHSC, CEHSC+Trace and NHSC+Cent are three hypergraph spectral clustering methods. The only difference between NHSC and NHSC+Cent is the weights of hyperedges. NHSC uses the mean of distances between each two samples in a hyperedge while NHSC+Cent uses the mean of distances between the hyperedge centroid and the samples in a hyperedge. NHSC+Cent and CEHSC+Trace are referenced from our recent work~\cite{hyperweight} which empirically studied the effect of hyperedge weighting scheme to hypergraph learning and select the most optimal weighting schemes for different hypergraph frameworks. NHSC+Cent and CEHSC+Trace are the best hyperedge weighting scheme and hypergraph spectral clustering combinations for addressing clustering issue reported in~\cite{hyperweight}.

We employ Sparse Representation-based Classifier (SRC)~\cite{src}, Collaborative Representation-based Classifier (CRC)~\cite{crc}, LIB-Support Vector Machine (LIBSVM)~\cite{libsvm}, Normalized Hypergraph Transduction (NHT)~\cite{hyper}, Graph Transduction (GT)~\cite{gt}, Adaptive Hypergraph-based Classifier (AHC)~\cite{adaptive}, Normalized Hypergraph Transduction using Matrix Trace Weights (NHT+Trace)~\cite{hyperweight}, Normalized Hypergraph Transduction using the Volume of Simplex Weights (NHT+Volume)~\cite{hyperweight}, and Sparse Graph-based Classifier (SGC)~\cite{sgc} as the compared approaches for image classification. GT and SGC are the graph transduction algorithms while AHC, NHT, CEHT+Trace and NHT+Volume are the hypergraph transduction algorithms. CEHT+Trace and NHT+Volume are the best hyperedge weighting scheme and hypergraph transduction combinations for addressing classification issue reported in~\cite{hyperweight}. In the experiments, all the compared methods are well tuned.

\begin{table*}[!tbp]
    \caption{Image Clustering performance comparison (in percents) on AR, ORL, COIL20, ETH80, Scene15 and Caltech256 databases. (In percentage)}
    \label{cluster}
    \vspace{-0.4cm}
\begin{center}
    \begin{tabular}{c | c c c c c c | c c c c c c}
    \hline
     \multirow{2}*{Methods}
    &\multicolumn{6}{c|}{Clustering Accuracy}&\multicolumn{6}{c}{Normalized Mutual Information (NMI)}\\ \cline{2-13}
    & AR& ORL& COIL20& ETH80 & Scene15 & Caltech256 & AR& ORL& COIL20& ETH80 & Scene15 & Caltech256  \\
    \hline
    NMF~\cite{nmf}&24.29&51.25&60.59&46.86&59.33&38.25    &56.12&70.31&70.89&41.13&58.41&38.89\\
    GRNMF~\cite{gnmf}&28.86&65.75&82.22&52.16&62.87&39.80   &60.52&82.19&89.99&46.93&61.32&38.45\\
    NCut~\cite{ncut}&61.79& 67.75&69.60&45.61&56.87 &38.00   &80.71&82.01&77.00&38.02&60.21&38.69\\
    NHSC~\cite{hyper}&36.71&66.75&76.60&50.85&58.87&36.40    &64.34&81.57&85.26&47.76&58.46&37.32 \\
    LSC~\cite{lsc}&35.86&66.00&76.04&55.55&66.33&43.95   &65.43&82.39&86.13&56.22&64.01&41.32\\
    CEHSC+Trace~\cite{hyperweight}&35.50&70.50&82.29&51.89&67.47&36.30   &64.00&82.75&89.12&46.83&62.03&34.01\\
    NHSC+Cent~\cite{hyperweight}&36.71&69.50&76.60&46.86&63.33&44.15  &64.36&81.62&85.26&46.72&59.53&43.29\\
    $L_1$GSC~\cite{l1graph}&59.93&69.50&68.13&51.77&56.27&34.30   &78.77&82.50&77.76&50.00&56.74&37.86\\
    \hdashline
    \textbf{$\bm{L_1}$HSC}&70.21&78.25&\textbf{82.85}&\textbf{56.34}&67.60&42.15   &86.71&87.21&\textbf{90.99}&\textbf{58.72}&64.40&41.88\\
    \textbf{$\bm{L_2}$HSC}&\textbf{73.43}&\textbf{79.50}&82.15&53.38&\textbf{69.20}&\textbf{44.95}&\textbf{86.73}&\textbf{87.22}&89.26&53.97&\textbf{65.80}&\textbf{43.60}\\

      \hline
    \end{tabular}
\end{center}
\end{table*}

\subsection{Image Clustering}
We conduct the image clustering experiments on all six image databases. For each database, the cluster number is fixed to its category number. Following~\cite{gnmf,lpi}, Clustering Accuracy and Normalized Mutual Information (NMI) are leveraged as two evaluation metrics.

Table~\ref{cluster} reports the clustering performances of different clustering methods. The bold number indicates the best accuracy in a dataset under same metric. We can find that $L_1$HSC and $L_2$HSC outperform almost all compared approaches in all experiments and achieve remarkable improvements over the conventional hypergraph spectral clustering algorithms. For examples, the clustering accuracy gains of $L_1$HSC over NHSC are 33.50\%, 11.50\%, 6.25\%, 5.49\%, 9.73\% and 5.75\%  on AR, ORL, COIL20, ETH80, Scene15 and Caltech256 datasets respectively. Similarly, such gains of $L_2$HSC are 36.72\%, 11.75\%, 5.55\%, 10.33\% and 8.55\%. The experimental results also show that, as the generalization of sparse graph spectral clustering, $L_1$HSC and $L_2$HSC obtain the better performances over $L_1$GSC which is known as a sparse graph spectral clustering algorithm. More specifically, The NMI gains of $L_1$HSC over $L_1$GSC on AR, ORL, COIL20, ETH80, Scene15 and Caltech256 datasets are 7.94\%, 4.71\%, 13.23\%, 8.72\%, 7.66\% and 4.02\% respectively. Similarly, the NMI improvements of $L_2$HSC over $L_1$GSC are 7.96\%, 4.72\%, 11.50\%, 3.97\%, 9.06\% and 5.74\% on AR, ORL, COIL20, ETH80, Scene15 and Caltech256 datasets. Another interesting phenomenon can be observed from the experimental results is that $L_2$HSC performs much better than $L_1$HSC comprehensively. We attribute this to the fact that Collaborative Representation (CR) is often more discriminative than Sparse Representation (SR)~\cite{crc,rcr}.

\begin{table}[h]
\footnotesize
\begin{center}
    \caption{Classification performance comparison (in percents) using AR, COIL20, ETH80 and Scene15 databases.}
    \label{imgcls}
    \begin{tabular}{p{1.8cm}<{\centering}|p{1.2cm}<{\centering} p{1.2cm}<{\centering} p{1.2cm}<{\centering} p{1.2cm}<{\centering}}
    \hline
     \multirow{2}*{Methods}
    &\multicolumn{4}{c}{Classification Errors (Mean~$\pm$~Standard Deviation, \%)}\\ \cline{2-5}
    &AR  & COIL20 & ETH80 & Scene15 \\
    \hline
    CRC~\cite{crc}&25.07$\pm$1.31&11.19$\pm$1.10&20.34$\pm$6.34&26.73$\pm$1.43\\
    SRC~\cite{src}&19.57$\pm$2.22&11.81$\pm$0.98&29.70$\pm$8.80&26.80$\pm$2.83\\
    LIBSVM~\cite{libsvm}&26.50$\pm$0.10&12.50$\pm$1.18&30.82$\pm$5.99&25.40$\pm$2.17\\
    NHT~\cite{hyper}&32.00$\pm$1.41&4.86$\pm$0.00&27.59$\pm2.98$&26.80$\pm$1.89\\
    GT~\cite{gt}&29.57$\pm$0.81&43.13$\pm$0.49&27.32$\pm$0.95&32.40$\pm$2.07\\
    AHC~\cite{adaptive}&33.14$\pm$1.21&10.13$\pm$1.41&27.13$\pm$2.85&25.54$\pm$1.59\\
    NHT+Trace~\cite{hyperweight}&32.21$\pm$1.31&3.68$\pm$0.69&26.10$\pm$4.31&25.47$\pm$1.89\\
    NHT+Volume~\cite{hyperweight}&32.29$\pm$1.21&4.79$\pm$0.29&25.82$\pm$4.61&25.40$\pm$1.98\\
SGC~\cite{sgc}&16.79$\pm$1.91&9.44$\pm$1.18&28.38$\pm$6.25&26.13$\pm$0.42\\
    \hdashline
\textbf{$\bm{L_1}$HT}&10.50$\pm$1.91&\textbf{2.15$\pm$0.49}&17.44$\pm$1.98&26.13$\pm$3.96\\
\textbf{$\bm{L_2}$HT}&\textbf{4.79$\pm$0.10}&4.51$\pm$4.42&\textbf{17.32$\pm$0.43}&\textbf{25.27$\pm$2.92}\\
  \hline
    \end{tabular}
\end{center}
\end{table}

\subsection{Image Classification}
We employ AR, COIL20, ETH80 and Scene15 databases for evaluating image classification performances of different classifiers. The two-fold cross-validation scheme is applied in the image classification experiments.

Table~\ref{imgcls} reports the classification errors of different classifiers on different databases. Similar to the experimental results of image clustering, the proposed approaches, $L_1$HT and $L_2$HT achieve very promising classification performance and consistently outperform the other hypergraph transduction approaches. Particularly, It is worthwhile to point out that $L_2$HT ranks first on three databases among all four databases. More specifically, the classification accuracy gains of $L_2$HT over the other hypergraph transduction approaches, namely NHT, AHC, NHT+Trace and NHT+Volume on AR database are 27.21\%, 28.35\%, 27.42\% and 27.50\% respectively. Such numbers of $L_1$HT on ETH80 database are 10.15\%, 9.69\%, 8.66\% and 8.50\% respectively. From the observations, it is not hard to find that $L_1$HT often performs much better than the SGC which can be deemed as the pairwise version of $L_1$HT. We believe that such phenomenon well verifies the better representational power of $L_1$-hypergraph over $L_1$-graph.

\begin{figure}[!t]
\centering
\subfigure[Noise Level = 0.1]{
\centering
\includegraphics[scale=0.56]{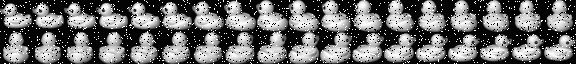}
}
\subfigure[Noise Level = 0.2]{
\centering
\includegraphics[scale=0.56]{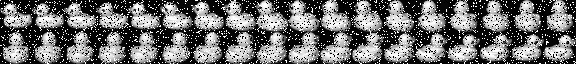}
}
\subfigure[Noise Level = 0.3]{
\centering
\includegraphics[scale=0.56]{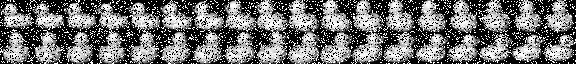}
}
\subfigure[Noise Level = 0.4]{
\centering
\includegraphics[scale=0.56]{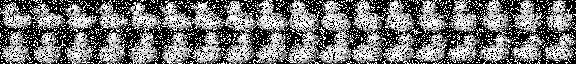}
}
\subfigure[Noise Level = 0.5]{
\centering
\includegraphics[scale=0.56]{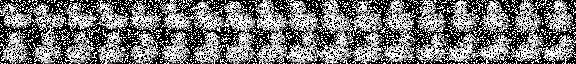}
}
\caption{The samples from five noisy versions of COIL20 database.}
\label{noise}
\end{figure}

\begin{figure*}[!t]
\centering
\subfigure[Accuracy (Image Clustering) $\bm{\uparrow}$]
{\label{noisea}
\centering
\includegraphics[scale=0.28]{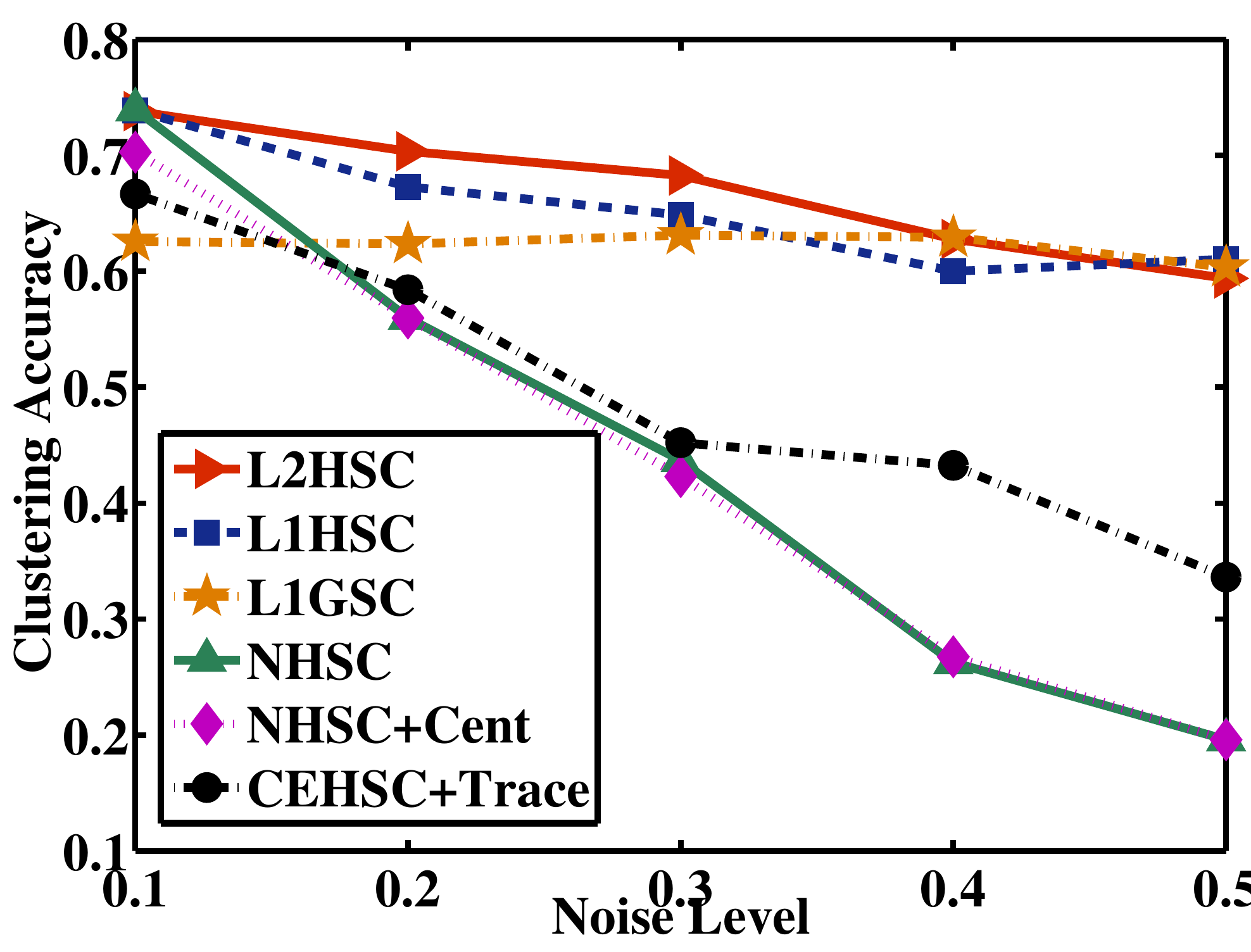}
}
\subfigure[NMI (Image Clustering) $\bm{\uparrow}$]{
\label{noiseb}
\centering
\includegraphics[scale=0.27]{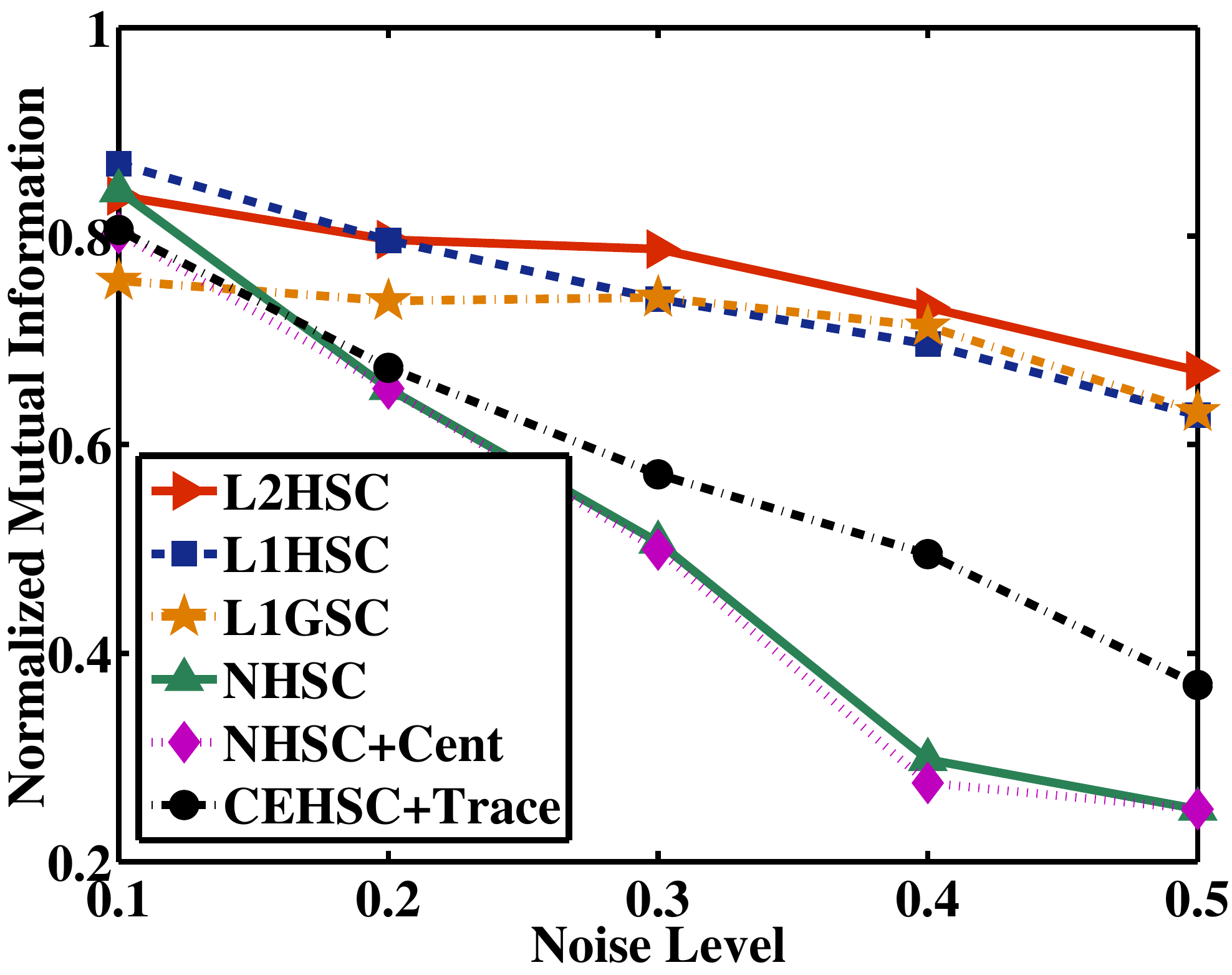}
}
\subfigure[Error (Image Classification) $\bm{\downarrow}$]{
\label{noisec}
\centering
\includegraphics[scale=0.28]{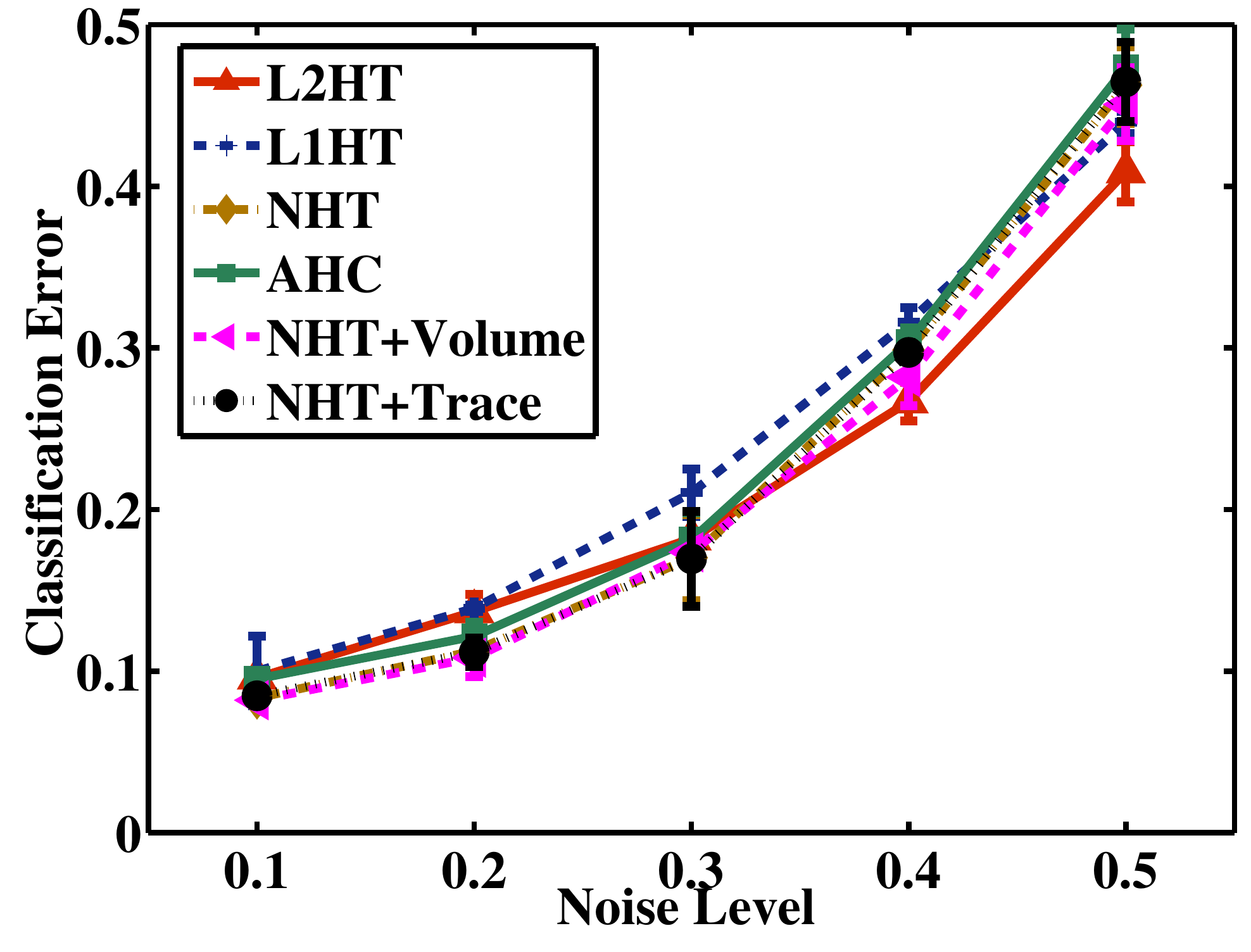}
}
\caption{The image clustering or classification performances of different hypergraph-based approaches using different noisy versions of COIL20 database. $\bm{\uparrow}$ means the higher the better while $\bm{\downarrow}$ means the lower the better.}
\label{noiseperformance}
\end{figure*}

\subsection{Robustness Analysis}
It is well known that Sparse Representation (SR) and Collaborative Representation (CR) are robust to the disguise and noise. Since $L_1$-Hypergraph ($L_1$H) and $L_2$-Hypergraph ($L_2$H) are generated by SR and CR, here we conduct several experiments to see if $L_1$H and $L_2$H also enjoy such desirable properties.

\subsubsection{Robust to Noise}
In order to evaluate the robustness of our works to noise, we construct five noisy versions of COIL dataset via randomly assigning zero or 255 to a certain proportion of pixels in the image. In these noisy COIL20 databases, we use the \textbf{Noise Level} to measure the noise degree of data where the \textbf{Noise Level} is defined as the ratio of the number of noisy dimensions to the number of total dimensions. The \textbf{Noise Levels} of these five noisy versions of COIL20 database are 0.1, 0.2, 0.3, 0.4 and 0.5 respectively (see the examples in Figure~\ref{noise}). We follow the same fashions as mentioned in the previous sections to conduct the image classification and clustering experiments in these noisy image databases.

We plot the experimental results of different approaches under different Noise Levels in Figure~\ref{noiseperformance}. Figures~\ref{noisea} and~\ref{noiseb} report the clustering accuracy and NMI respectively. From these observations, we can see that the clustering performances of all approaches are decreased along with the increasing of Noise Levels. However, compared with the conventional hypergraph methods, such as NHSC and CEHSC,  $L_1$GSC, $L_1$HSC and $L_2$HSC perform much better and their clustering performances are apparently decreased much slower along with the increasing of Noise Level. Clearly, such phenomenon shows that $L_1$HSC and $L_2$HSC are more robust to noise in comparison with the conventional hypergraph spectral clustering models. Figure~\ref{noisec} shows the classification errors of different approaches under different Noise Levels. The observations of image classification experiments are quite different to the ones of the aforementioned image clustering experiments. The performances of all approaches are very similar, and only $L_2$HT slightly outperforms the others. We mainly attribute this phenomenon to the fact that the training samples all contain noise and the supervision from category labels compulsively introduces the noise samples to the hypergraph learning models. Therefore, the contributions of the robust sample selection procedures from Collaborative Representation (CR) and Sparse Representation (SR) are weakened. Comprehensively speaking, $L_1$H and $L_2$H enjoy a certain amount of robustness to noise particularly in the unsupervised way.

\begin{table}[!tbp]
    \caption{Classification performance comparison (in percents) on AR database with real disguise.}
    \label{clusterdisg}
    \vspace{-0.2cm}
\begin{center}
    \begin{tabular}{c | p{0.9cm}<{\centering} p{1.2cm}<{\centering} p{1.25cm}<{\centering} p{1.2cm}<{\centering} p{0.9cm}<{\centering} }
    \hline
     \multirow{2}*{Metric}
    &\multicolumn{5}{c}{Methods}\\ \cline{2-6}
    & NHSC&  NHSC+Cent& CEHSC+Trace& \textbf{$\bm{L_1}$HSC} & \textbf{$\bm{L_2}$HSC}\\
     \hline
    AC&20.32&20.38&19.06&48.00&68.19\\
    NMI&46.11&45.99&46.60&67.81&80.42\\
    \hline
    \end{tabular}
\end{center}
\end{table}

\begin{table}[!tbp]
\begin{center}
    \caption{Classification performance comparison (in percents) using AR, COIL20, ETH80 and Scene15 databases.}
    \label{classdisg}
    \begin{tabular}{c|c}
    \hline
   Methods&Classification Errors (\%)\\
    \hline
    CRC~\cite{crc}&17.08\\
    SRC~\cite{src}&58.25\\
    NHT~\cite{hyper}&92.67\\
    AHC~\cite{adaptive}&92.83\\
    NHT+Trace~\cite{hyperweight}&92.67 \\
    NHT+Volume~\cite{hyperweight}&92.92\\
SGC~\cite{sgc}&20.00\\
    \hdashline
\textbf{$\bm{L_1}$HT}&20.42\\
\textbf{$\bm{L_2}$HT}&\textbf{4.00}\\
  \hline
    \end{tabular}
\end{center}
\end{table}

\subsubsection{Robust to Disguise}
We employ AR database to evaluate the image clustering and classification performances under the disguise case, since AR database have already provided the face images with natural disguise (sunglasses and scarfs) for each subject. In the image clustering experiments, all the face images are applied. In the image classification case, we use the face images without any occlusion for training while use the face images with occlusion for testing.

Table~\ref{clusterdisg} lists the clustering performances of our works and three other conventional hypergraph approaches. In these experiments, $L_1$HSC and $L_2$HSC show the significant advantages. For examples, the clustering accuracy gains of $L_1$HSC over NHSC, NHSC+Cent and CEHSC+Trace are 27.68\%, 27.62\% and 29.00\% respectively. $L_2$HSC performs even better. The clustering accuracy improvements of $L_2$HSC over these conventional hypergraph approaches are 47.87\%, 47.81\% and 49.13\% respectively. Table~\ref{classdisg} tabulates the classification errors of different approaches in the disguise case. From the observations, $L_1$HT and $L_2$HT obtain very promising performance while all the other traditional hypergraph transduction approaches are totally failed. The classification error of $L_2$HT is 23 times lower than the ones of NHT, AHC, NHT+Trace and NHT+Volume. Clearly, the experimental results demonstrate that $L_1$HT and $L_2$HT are robust to disguise.

\begin{figure}[!tb]
\centering
\subfigure[The impact of $\beta$ to $L_1$HSC]{
\centering
\includegraphics[scale=0.2]{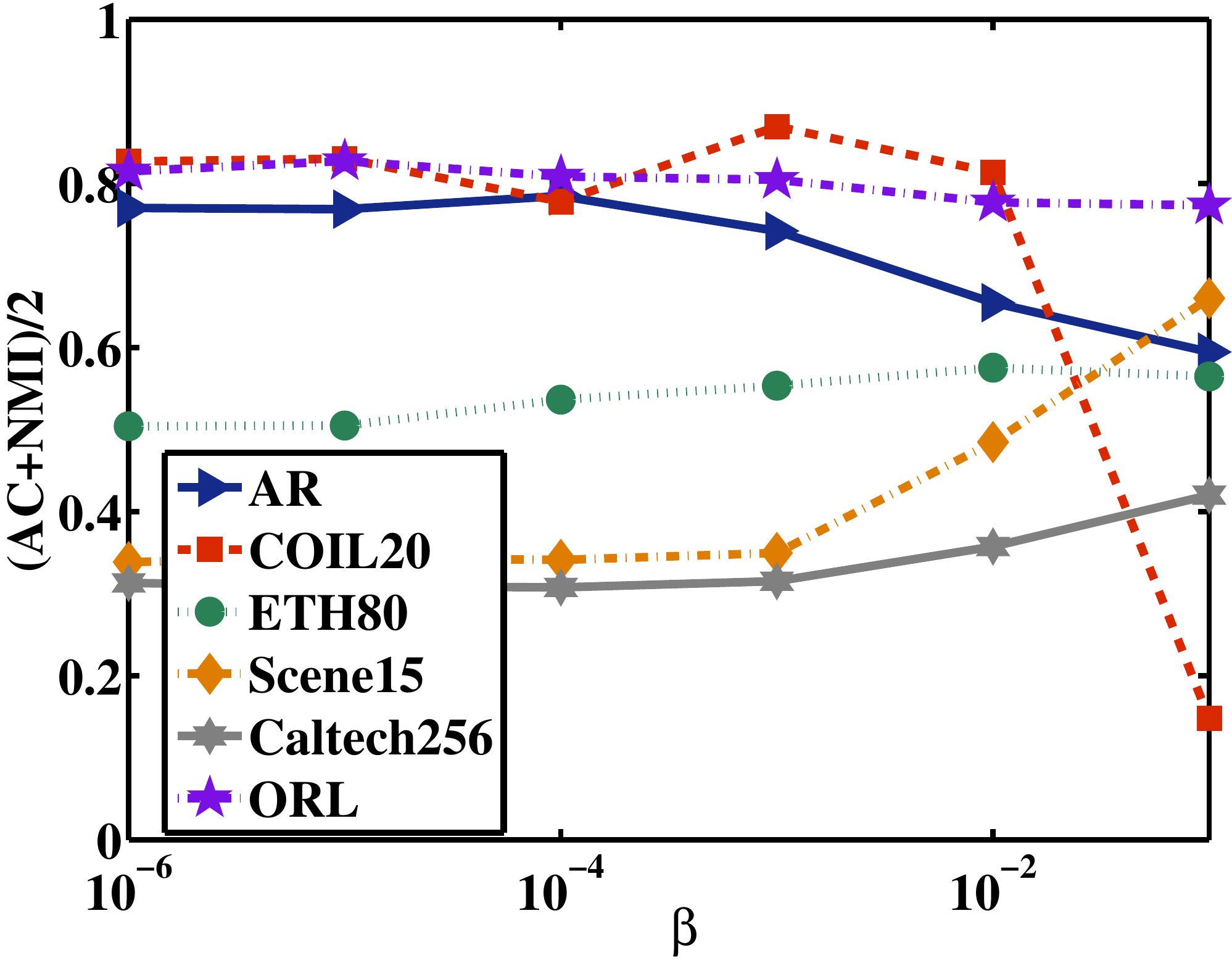}
}
\subfigure[The impact of $\beta$ to $L_2$HSC]{
\centering
\includegraphics[scale=0.2]{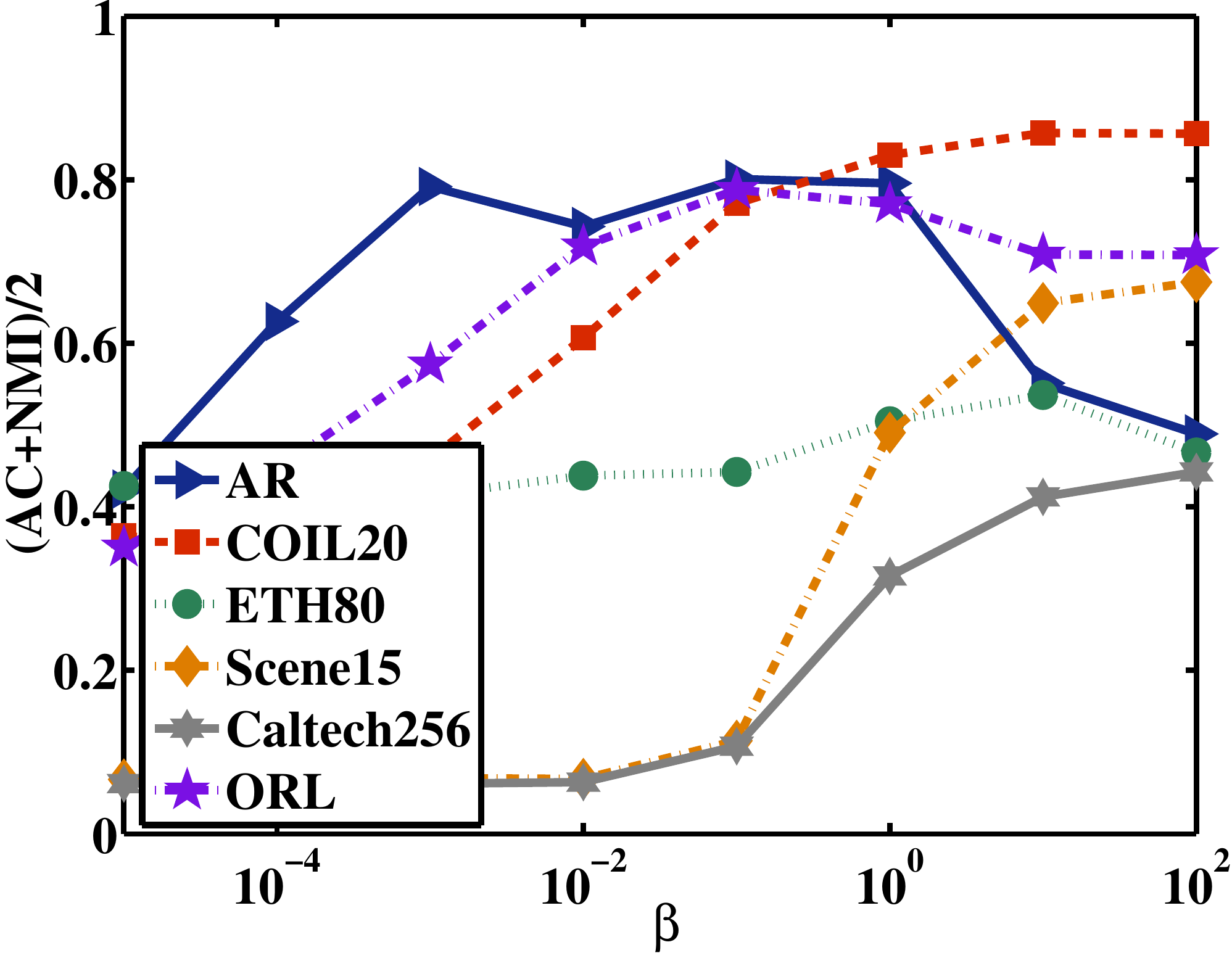}
}
\caption{The impacts of different parameters to the clustering performances of $L_1$HSC and $L_2$HSC.}
\label{clusterparam}
\end{figure}

\begin{table}[!tbp]
    \caption{The hyperedge length ($t$) candidate collections of $L_1$HSC and $L_2$HSC and their optimal hyperedge lengths in different datasets.}
    \label{rhsct}
    \vspace{-0.2cm}
\begin{center}
    \begin{tabular}{c | c| c c }
    \hline
     \multirow{2}*{Database}&\multirow{2}*{Candidates}
    &\multicolumn{2}{c}{The optimal $t$}\\ \cline{3-4}
     &&$L_1$HSC & $L_2$HSC\\
     \hline
  ORL&\{2:11\}&$t=9$&$t=11$\\
  AR& \{2:14\}&$t=13$&$t=14$\\
  COIL20& \{3,5,10,20,36,54,72\}&$t=3$&$t=5$\\
  ETH80& \{3,5,10,41:41:410\}&$t=10$&$t=123$\\
   Scene15&\{5:5:50,60:10:100\}&$t=50$&$t=100$\\
  Caltech256&\{5:5:50,60:10:100\}&$t=100$&$t=100$\\
    \hline
    \end{tabular}
\end{center}
\end{table}

\begin{figure*}[!tb]
\centering
\subfigure[The impact of $\beta$ to $L_1$HT]{
\centering
\includegraphics[scale=0.205]{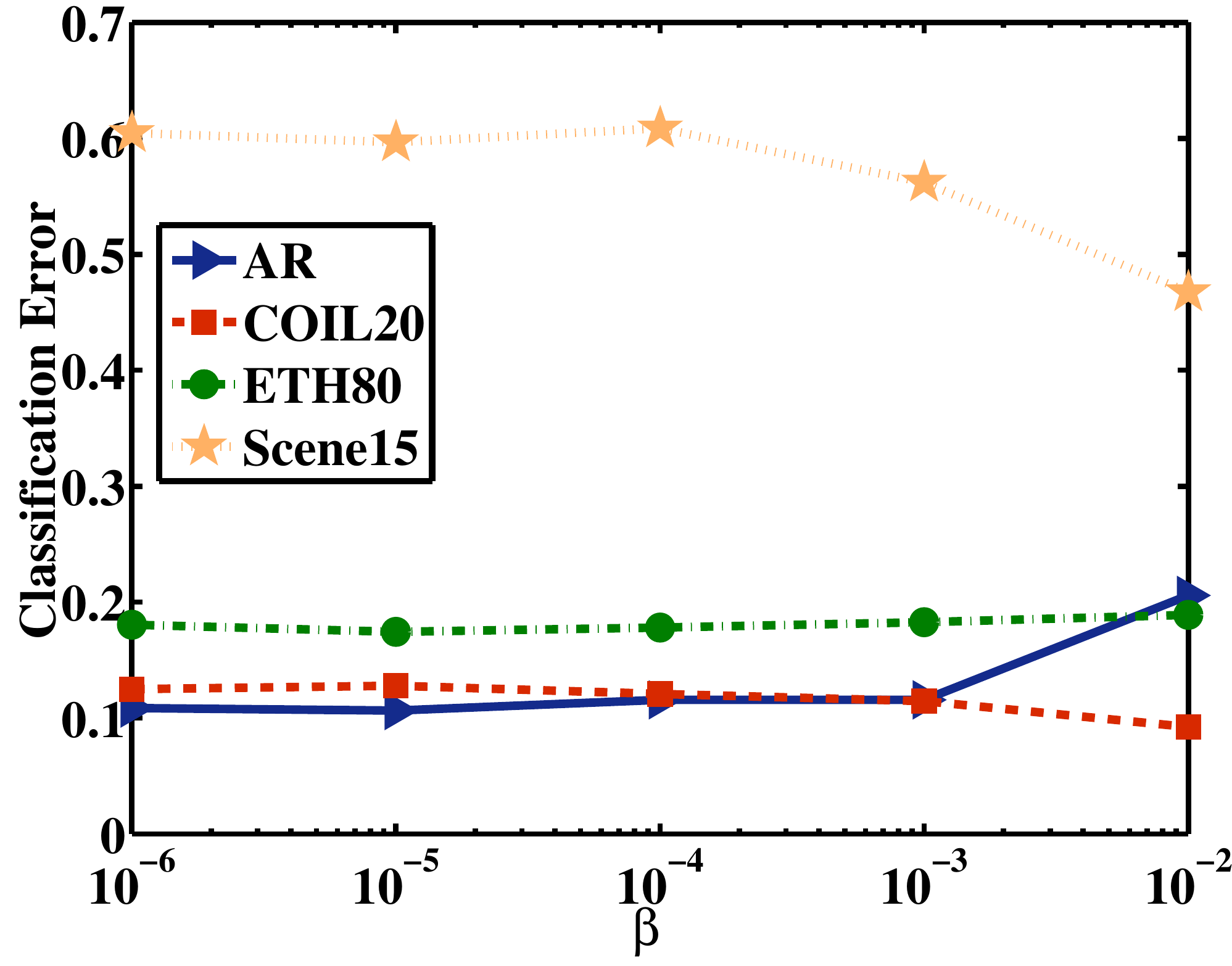}
}
\subfigure[The impact of $\beta$ to $L_2$HT]{
\centering
\includegraphics[scale=0.205]{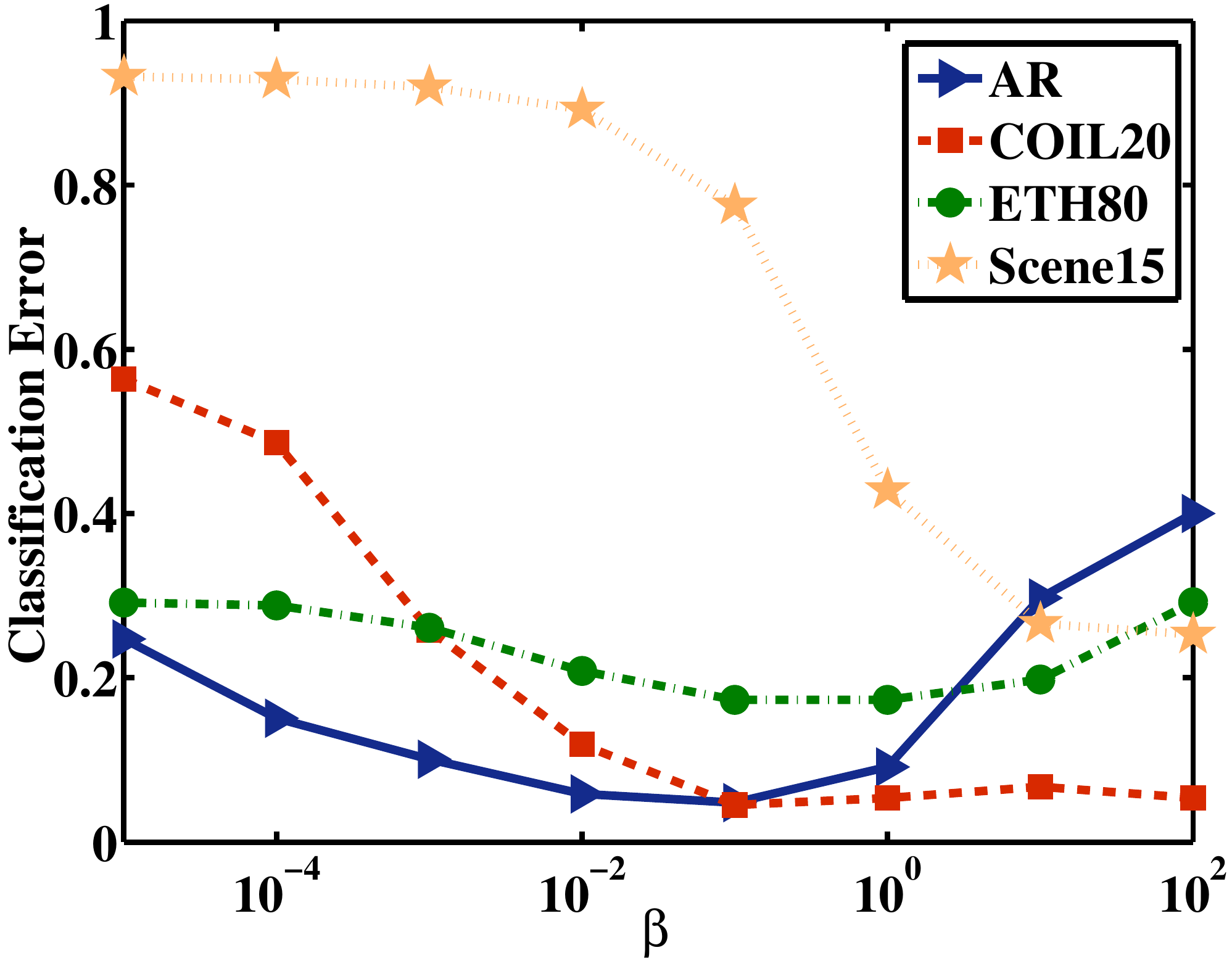}
}
\subfigure[The impact of $\lambda$ to $L_1$HT]{\label{laml1ht}
\centering
\includegraphics[scale=0.205]{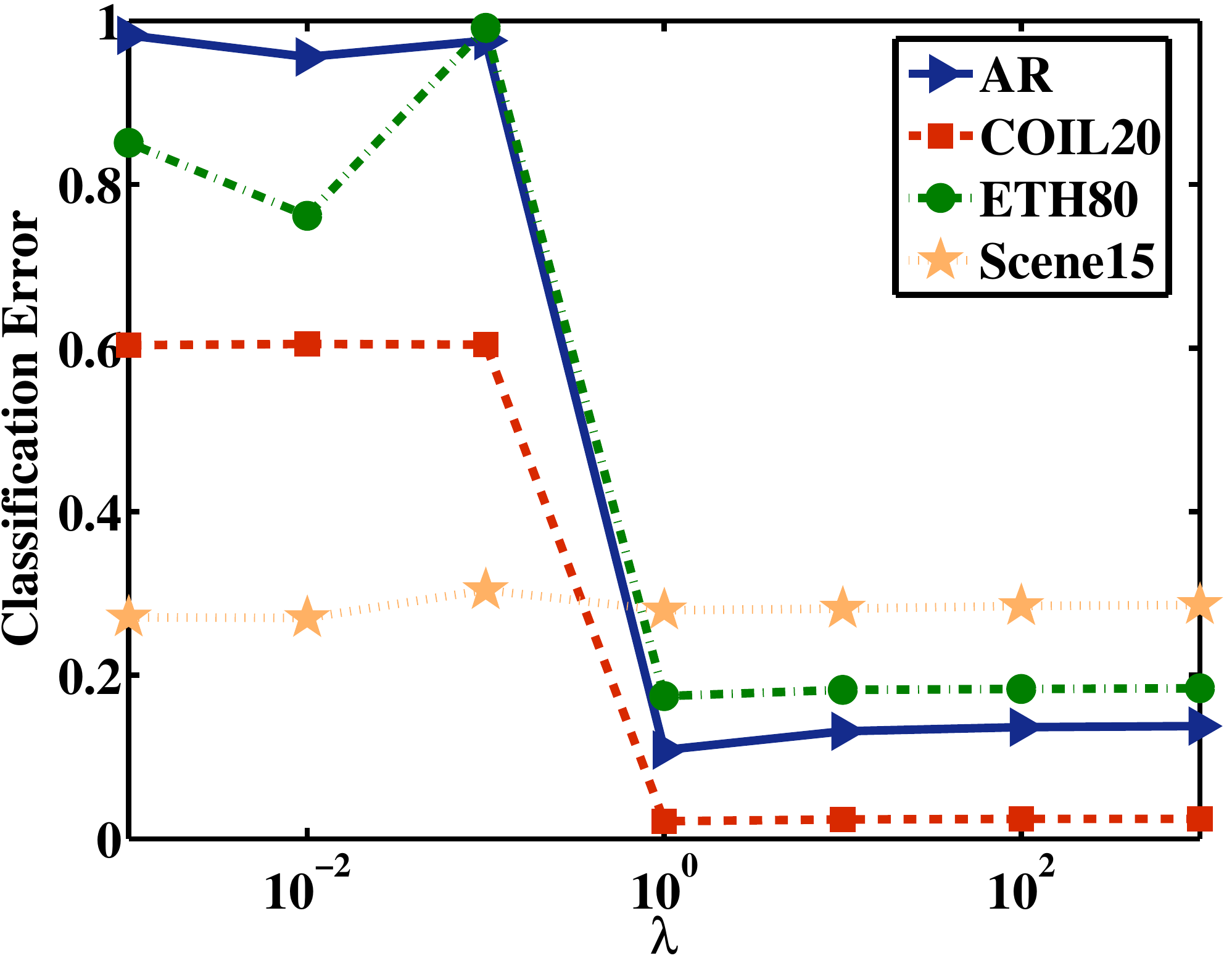}
}
\subfigure[The impact of $\lambda$ to $L_2$HT]{\label{laml2ht}
\centering
\includegraphics[scale=0.205]{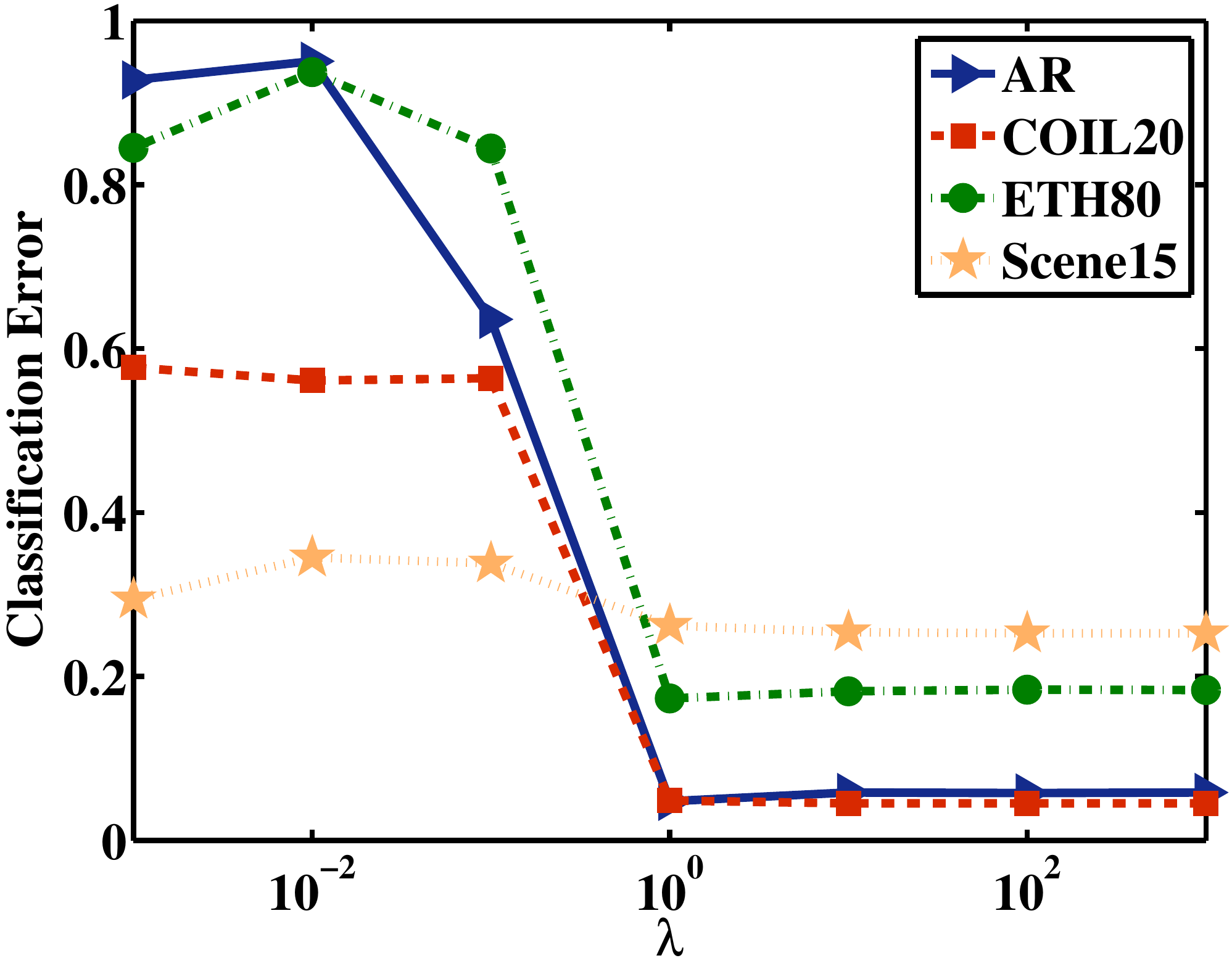}
}
\caption{The impacts of different parameters to the classification performances of $L_1$HT and $L_2$HT.}
\label{clsparam}
\end{figure*}
\begin{table}[!tbp]
    \caption{The hyperedge length ($t$) candidate collections of $L_1$HT and $L_2$HT and their optimal hyperedge lengths in different datasets.}
    \label{rhtt}
    \vspace{-0.2cm}
\begin{center}
    \begin{tabular}{c | c |c c }
    \hline
     \multirow{2}*{Database}&\multirow{2}*{Candidates}
    &\multicolumn{2}{c}{The optimal $t$}\\ \cline{3-4}
     &&$L_1$HT & $L_2$HT\\
     \hline
  AR& \{2:14\}&$t=5$&$t=8$\\
  COIL20& \{3,5,10,20,36,54,72\}&$t=5$&$t=3$\\
  ETH80& \{3,5,10,41:41:410\}&$t=10$&$t=10$\\
   Scene15&\{5:5:50,60:10:100\}&$t=20$&$t=35$\\
    \hline
    \end{tabular}
\end{center}
\end{table}

\subsection{Parameter Settings}
In this section, we discuss the influences of different parameters to our works. The RHSC methods, such as $L_1$HSC and $L_2$HSC, mainly involve two parameters, $\beta$ and $t$, where $\beta$ is used for controlling the regression regularization term in a regression model while $t$ is the hyperedge length. The RHT methods, such as $L_1$HT and $L_2$HT, have one more parameter, $\lambda$, which is used for balancing the labeling error of data and the loss of hypergraph partition. Although the hyperedge length $t$ is an important parameter which can deeply influence the quality of hypergraph, it has numerous possible values and it is impossible for us to evaluate each possible value. So, in our experiments, we define a hyperedge length candidate collection and then conduct experiments to empirically find the optimal value of hyperedge length. Tables~\ref{rhsct} and~\ref{rhtt} respectively list the hyperedge length candidate collections of the proposed RH works and report their optimal hyperedge lengths. Figure~\ref{clusterparam} demonstrates the influences of $\beta$ to the clustering performances of $L_1$HSC and $L_2$HSC. We choose the mean of clustering accuracy and NMI as the comprehensive clustering performance evaluation metrics. From the observations in Figure~\ref{clusterparam}, we can find that the performance of $L_1$HSC is quite insensitive to $\beta$ and the optimal $beta$ of $L_1$HSC on ORL, AR, COIL20, ETH80, Scene15 and Caltech256 databases, are $10^{-5}$, $10^{-4}$, $10^{-3}$, $10^{-2}$ and $10^{-1}$ respectively. With regard to $L_2$HSC, a higher value of $\beta$ is much better on Scene15, Caltech256, COIL20 and ETH80 databases while a medium $\beta$ is more suitable to the AR and ORL databases. Figure~\ref{clsparam} shows the impacts of $\beta$ and $\lambda$ to the classification performances of $L_1$HT and $L_2$HT. The phenomena as similar as the ones in Figure~\ref{clusterparam} are observed. $L_1$HSC is not very sensitive to $\beta$ and the best $\beta$ for AR, COIL20, ETH80 and Scene15 databases are $10^{-5}$, $10^{-3}$,  $10^{-5}$ and $10^{-2}$ respectively. A medium $\beta$ works well for $L_2$HT on AR and COIL20 databases while $L_2$HT with a higher $\beta$ shows the better performances on Scene15 and ETH80 databases. Moreover, from the observations in Figures~\ref{laml1ht} and~\ref{laml2ht}, both $L_1$HT and $L_2$HT can achieve the best classification performances when $\lambda$ is larger than 1.

\section{Conclusion}
\label{conclude}
In this paper, we presented a new solution for hypergraph construction in which the regression models are used for measuring the closeness among samples. We named this new hypergraph framework Regression-based Hypergraph (RH). Based on two conventional hypergraph learning models, namely Hypergraph Spectral Clustering (HSC) and Hypergraph Transduction (HT), We also developed the Regression-based Hypergraph Spectral Clustering (RHSC) and Regression-based Hypergraph Transduction (RHT) models for addressing the clustering and classification issues. As two influential regression approaches for visual learning, Sparse Representation (SR) and Collaborative Representation (CR) are employed to instantiate two instances of RH and their RHSC and RHT algorithms. Six popular image databases are leveraged for validating the effectiveness of our works. It can be concluded from observations of the experiments that RH inherits the desirable properties from both hypergraph and regression model.

There still exist many interesting future works based on our models, since RH is a general framework for hypergraph construction in which researchers can flexibly choose their own appropriate regression models to construct RHs for tackling different tasks. For examples, RH can be further applied to hypergraph-based subspace learning~\cite{dhlp}, feature selection~\cite{ufs}, multi-label learning~\cite{hypersvm,mhyper} or attribute learning~\cite{hap}. Moreover, to investigate the adaptive and efficient RH construction fashion is also a very meaningful direction for improving the utility and efficiency~\cite{adaptive,largeg}.

\section*{Acknowledgement}
\bibliographystyle{IEEEtran}
\bibliography{mybib}

\end{document}